\documentclass{article}

\usepackage{microtype}
\usepackage{graphicx}
\usepackage{subfigure}
\usepackage{booktabs} 

\usepackage{hyperref}

\usepackage[accepted]{icml2024}

\usepackage{amsmath}
\usepackage{amssymb}
\usepackage{bbding}
\usepackage{mathtools}
\usepackage{amsthm}
\usepackage{multirow}
\usepackage{makecell}
\usepackage{enumitem}

\usepackage[capitalize,noabbrev]{cleveref}

\theoremstyle{plain}

\theoremstyle{definition}

\theoremstyle{remark}

\usepackage{bm}
\usepackage{graphicx}
\usepackage{graphics} 
\usepackage{subcaption}
\usepackage{colortbl}
\usepackage[textsize=tiny]{todonotes}

\icmltitlerunning{Unlocking the Power of Spatial and Temporal Information in Medical Multimodal Pre-training}
\begin{document}

\twocolumn[
\icmltitle{Unlocking the Power of Spatial and Temporal Information \\ in Medical Multimodal Pre-training}



\icmlsetsymbol{equal}{*}

\begin{icmlauthorlist}
\icmlauthor{Jinxia Yang}{sch}
\icmlauthor{Bing Su}{sch,bigdata}
\icmlauthor{Wayne Xin Zhao}{sch,bigdata}
\icmlauthor{Ji-Rong Wen}{sch,bigdata,Information}
\end{icmlauthorlist}

\icmlaffiliation{sch}{Gaoling School of Artificial Intelligence, Renmin University of China}
\icmlaffiliation{bigdata}{Beijing Key Laboratory of Big Data Management and Analysis Methods}
\icmlaffiliation{Information}{School of Information, Renmin University of China}

\icmlcorrespondingauthor{Bing Su}{subingats@gmail.com}
\icmlcorrespondingauthor{Wayne Xin Zhao}{batmanfly@gmail.com}

\icmlkeywords{Machine Learning, ICML}

\vskip 0.3in
]



\printAffiliationsAndNotice{}  

\begin{abstract}
Medical vision-language pre-training methods mainly leverage the correspondence between paired medical images and radiological reports. Although multi-view spatial images and temporal sequences of image-report pairs are available in off-the-shelf multi-modal medical datasets, most existing methods have not thoroughly tapped into such extensive supervision signals. In this paper, we introduce the Med-ST framework for fine-grained spatial and temporal modeling to exploit information from multiple spatial views of chest radiographs and temporal historical records.  For spatial modeling, Med-ST employs the \emph{Mixture of View Expert~(MoVE)} architecture to integrate different visual features from both frontal and lateral views. 
To achieve a more comprehensive alignment, Med-ST not only establishes the global alignment between whole images and texts but also introduces modality-weighted local alignment between text tokens and spatial regions of images.
For temporal modeling, we propose a novel cross-modal bidirectional cycle consistency objective by forward mapping classification~(FMC) and reverse mapping regression~(RMR). By perceiving temporal information from simple to complex, Med-ST can learn temporal semantics. 
Experimental results across four distinct  tasks demonstrate the effectiveness of Med-ST, especially in temporal classification tasks. Our code and model are available at \url{https://github.com/SVT-Yang/MedST}. 

\end{abstract}    
\section{Introduction}
\label{sec:intro}

\begin{figure}[!t]
	\begin{center}
		\includegraphics[width=\linewidth]{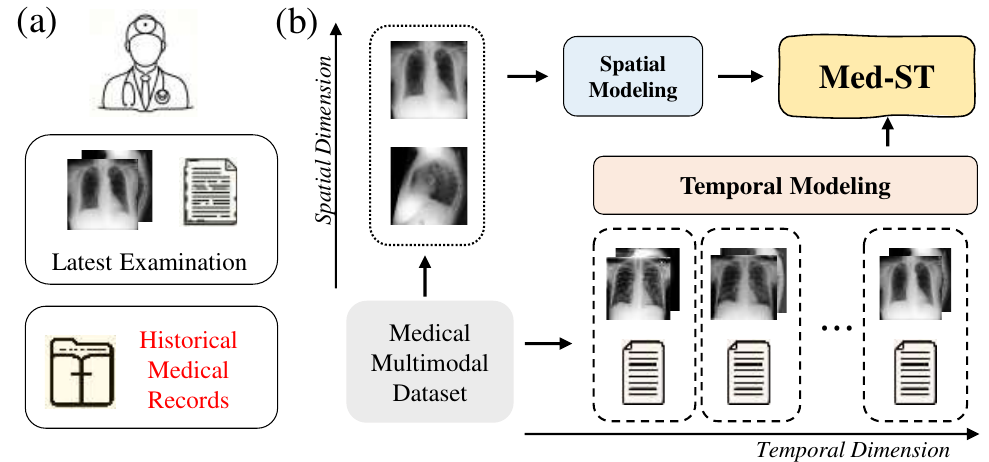}
	\end{center}
	\caption{The motivation and our framework: (a) The practice of physicians in a clinical setting. (b) We propose the Med-ST framework, which explicitly designs spatial and temporal modeling by mining spatio-temporal supervision signals from the dataset.} 
 \label{fig:motivation}
\end{figure}
Medical vision-language pre-training aims to learn visual and textual representations from paired radiological reports and images. The learned representations can be adapted to various downstream tasks, which have great potential value for clinical auxiliary diagnosis and so on. Existing medical vision language pre-training methods generally focus on utilizing the correspondences between paired images and texts~\cite{huang2021gloria, wang2022multi, zhou2022generalized,wan2023med, zhou2023advancing}. 
The radiological report is aligned with the  frontal medical image through contrastive learning or masked modality reconstruction~\cite{cheng2023prior, zhou2023advancing, moon2022multi}. Although the learned visual representations can encode medical semantic information, they cannot fully capture fine-grained spatial information in different views, nor can they distinguish fine-grained temporal differences between image-text pairs of the same patient at different times. 

The applicability of medical vision-language pre-training models significantly depends on their capacity to mimic human cognitive process in interpreting radiological reports and images.
Typically, in a clinical setting as shown in Figure~\ref{fig:motivation}(a),  the doctor's diagnosis relies on the comprehensive examination results, considering frontal views and lateral views. This process also involves reviewing the patient's historical medical records to comprehend past symptoms and trends. 
Fortunately, such comprehensive spatio-temporal information is  available in off-the-shelf medical multimodal datasets. Besides image-text pairs, these datasets naturally embody parts of two types of supervision signals: spatial information (frontal and lateral views) and temporal information (chronological order of diagnoses). 

However, these supervision signals have not been fully explored by  existing medical pre-training methods. For spatial information, images of lateral views are either ignored or treated identically to those of frontal views ~\cite{huang2021gloria, wang2022multi,zhou2023advancing, dawidowicz2023limitr, shu2024miter}. 
Regarding temporal information, existing work only utilizes a single prior image and does not design objectives specifically for this self-supervised signal~\cite{bannur2023learning}. The temporal information in  historical sequences of image-report pairs has not yet been explored. 

In this work, we introduce the Med-ST framework, which jointly exploits the comprehensive spatial and temporal information within off-the-shelf medical datasets to supervise the pre-training of visual and textual representations. As visualized in Fig.~\ref{fig:motivation}(b), Med-ST comprises two main components: spatial modeling and temporal modeling. In spatial modeling, we introduce the \emph{Mixture of View Expert (MoVE)} architecture to construct the multi-view image encoder. We feed both frontal and lateral views into the encoder, which utilizes two experts to extract complementary information from different spatial perspectives. Features generated by both experts are integrated to form a joint visual representation of these varied spatial angles. Visual representations are then aligned with textual representations through contrastive learning~\cite{oord2018representation}. Recognizing that pathological areas occupy only a part of the images or texts, we propose modality-weighted local alignment, which assigns different weights to different local image patches and text tokens pairs based on the information they contain, achieving a fine-grained local alignment.
For temporal modeling, we encourage the learned image-text feature sequences to express the same semantic changes, allowing the pre-training model to gain more supervision signals. Motivated by~\citet{dwibedi2019temporal},  we perform 
bidirectional cycle consistency between sequences of different modalities.
Since the model has never explicitly perceived temporal sequence information, directly learning such information is challenging. Therefore, we adopt a novel bidirectional learning approach from simple to complex. In the forward process, we design a relatively simple classification loss to initially perceive information in the sequence. In the reverse process, we add a Gaussian prior for regression. We penalize inconsistencies by transforming the mismatch measurements into classification and regression targets. Through a bidirectional process from simple to complex, our model perceives the information of the sequence context, thus being able to capture changes. 
By incorporating spatial and temporal modeling, the global multi-view visual representation and textual representation interact in both  spatial and temporal dimensions. In this way, our Med-ST gains the ability to jointly encode fine-grained spatiotemporal information into multi-modal representations.

The contributions of our study are outlined as follows:
\begin{itemize}[topsep=0pt]
\setlength{\itemsep}{-4pt}
\item  We thoroughly explore the information in medical multimodal datasets without additional manual labeling. Beyond text-image pairing, we leverage multi-view spatial data and historical temporal data, yielding a richer set of supervision signals.
\item  Our spatial modeling utilizes the  \emph{MoVE} architecture to tackle both frontal and lateral views with specialized experts, and introduces  modality-weighted local alignment to establish fine-grained contrastive learning between spatial image regions and semantic tokens.
\item  For temporal modeling, we propose a novel cross-modal bidirectional cycle consistency objective that progresses from simple to complex. By forward mapping classification and reverse mapping regression, our model becomes capable of perceiving the context of sequences.
\item We evaluate the performance of our method in temporal tasks and medical image classification tasks. The results demonstrate the effectiveness of our method.
\end{itemize}

\section{Related Work}
\label{sec:related}
{\noindent\bf Vision-Language Pre-training Models.}
Multimodal vision-language pre-training~(VLP) is a field of research that aims to jointly learn representations of both visual and textual data. It  trains deep neural networks on large-scale datasets that contain both image and text data, mainly image-text pairs~\cite{radford2021learning, li2021align, li2023blip, wang2022image, alayrac2022flamingo,chen2022pali,bao2022vlmo, MIR-2022-07-221, MIR-2022-05-167}.
By learning to represent images and text in a shared embedding space, multimodal pre-training can generate cross-modal representations that capture the semantic meaning of both modalities.
Due to the impressive performance of VLP in the natural domain, it has also gained significant popularity in other fields, including the medical domain. It is important to note that directly transferring VLP models to the medical domain may not be feasible due to differences between the two domains~\cite{chambon2022roentgen, chambon2022adapting}.

{\noindent\bf Representation Learning in Medical Domain.}
The research on vision-language pre-training in medical domain can be divided into two branches: one is leveraging external knowledge bases~\cite{wu2023medklip, wang2022medclip, qin2022medical} and the other is  fully utilizing existing datasets through specific model design~\cite{huang2021gloria, wang2022multi, zhou2022generalized, zhou2023advancing, cheng2023prior}. In the first branch,  MedKLIP and MedCLIP~\cite{wu2023medklip, wang2022medclip} leverage entity extraction or descriptive information to enhance the model.  Currently, several large language models are being applied in  medical domain~\cite{wang2023large, singhal2022large, shaib2023summarizing, yunxiang2023chatdoctor}, primarily by using medical instruction datasets to generate more accurate answers. However, utilizing external knowledge may be limited by the availability and completeness of the external knowledge base. In the other branch,
MGCA~\cite{wang2022multi}  learns generalized medical visual representations through leveraging the inherent semantic correlations and MRM~\cite{zhou2023advancing} learns knowledge-enhanced semantic representations by reconstructing both masked image patches and masked report tokens.
However, these works only utilize the frontal view of the chest X-ray, discarding the lateral views~\cite{huang2021gloria, wang2022multi}, or treat them the same as frontal views without exploring the relationship between them~\cite{zhou2023advancing,dawidowicz2023limitr}. We believe  there is an association between the frontal and lateral views that could be helpful for downstream tasks, and medical evidence has shown the value of lateral views in diagnosis~\cite{raoof2012interpretation}. Therefore, we attempt to better utilize the lateral view through spatial modeling. 

{\noindent \bf Temporal Self-Supervised Learning in Medical Domain.}  
Effective representations can be learned from time series data through self-supervision, eliminating the need for manual annotation. Specifically, medical time series data often contain abundant semantic information.
BioViL-T~\cite{bannur2023learning} exploits time-related information by using prior images, thereby gaining additional supervision. However, BioViL-T only utilizes a single prior image and lacks a temporal optimization objective.  Our approach considers both text and image pairs in a temporal context and designs a time-related objective, i.e., cross-modality bidirectional cycle consistency, to better leverage the information provided by time series.


\begin{figure*}[t]
\begin{center}
\includegraphics[width=1.0\linewidth]{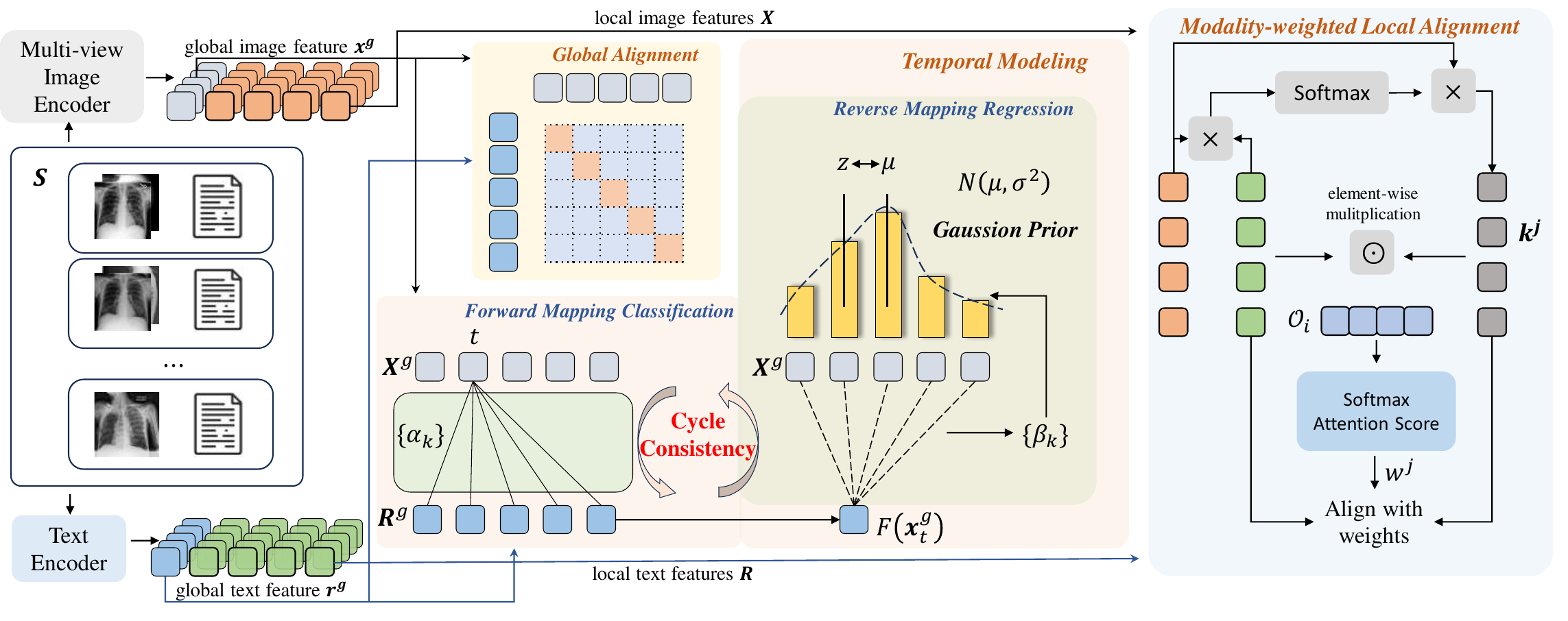}
\end{center}
\vskip -0.1 in
\caption{The framework of Med-ST. 
For spatial modeling, Med-ST extracts integrated multi-view visual representations, performs global alignment between paired global features, and introduces modality-weighted local alignment between paired local features. For temporal modeling, global image and text features in all pairs form two sequences respectively, and Med-ST imposes cross-modality bidirectional cycle consistency constraints between them.}
\label{fig:framework}
\end{figure*}
\section{Method}

\subsection{Overview}
\label{sec:overview}
Existing medical datasets contain some inherent spatial and temporal supervision signals that have been previously overlooked, obviating the need for supplementary manual annotations. A collection of images associated with a single report is referred to as a 
\emph{study}\footnote{https://physionet.org/content/mimic-cxr/2.0.0/}. Given a dataset comprising multiple \emph{studies}, each \emph{study} typically includes a frontal view image along with its corresponding radiological report. In some cases, lateral views may be present. All \emph{studies} of a patient have a chronological order. As a result, we can extract matched multi-view data and time-series data from the dataset, which provide spatial and temporal supervision signals for pre-training multi-modal representations. 

Formally, we reorganize the data in a medical dataset \( \mathcal{D} \) into a number of sequences, where each sequence contains a series of image-text pairs and typically corresponds to the historical diagnosis records of a patient. Some image-text pairs do not belong to any sequences, e.g., patients have only one diagnosis record. We regard such individual image-text pairs as sequences with a length of $1$. Therefore, \( \mathcal{D} = \{\bm{S}_1, \bm{S}_2, \cdots, \bm{S}_{|\mathcal{D}|}\} \), where $|\mathcal{D}|$ is the number of sequences in the dataset. For the $i$-th sequence, \( \bm{S}_i = \{(\bm{I}^f_{i,t}, \bm{I}^l_{i,t}, \bm{T}_{i,t})\}_{t=1}^{|\bm{S}_i|} \), where $\bm{I}^f_{i,t}$, $\bm{I}^l_{i,t}$, and $\bm{T}_{i,t}$ are the frontal image, the corresponding lateral image, and the related text at the $t$-th timestep, and $|\bm{S}_i|$ is the number of image-text pairs in this sequence. Not every image-text pair has a lateral view, i.e., for some $i$ and $t$, $\bm{I}^l_{i,t}$ does not exist, and we set it to a tensor with all zeros in this case.

Our proposed model Med-ST leverages both spatial and temporal modeling to learn multi-modal representations from such sequence data of multi-view image-text pairs. Med-ST consists of a multi-view image encoder $f_I$ and a text encoder $f_T$. The image encoder employs the proposed \emph{MoVE} architecture to extract integrated representations from frontal and lateral images, i.e., $[\bm{x}^g_{i,t}, \bm{X}_{i,t}]=f_I([\bm{I}^f_{i,t}, \bm{I}^l_{i,t}])$, where $\bm{x}^g_{i,t} \in \mathbb{R}^d$ is the global image feature, $\bm{X}_{i,t} \in \mathbb{R}^{M \times d}$ is the sequence of patch-wise features and $M$ is the number of tokens. The text encoder extracts the text representation, i.e., $[\bm{r}^g_{i,t}, \bm{R}_{i,t}]=f_T(\bm{T}^f_{i,t})$, where $\bm{r}^g_{i,t}  \in \mathbb{R}^d$ is the global text feature and $\bm{R}_{i,t}  \in \mathbb{R}^{W \times d}$ is the sequence of token-wise features and  $W$ is the number of tokens. As shown in Fig.~\ref{fig:framework}, in the training phase, for each input sequence $\bm{S}$, Med-ST first extracts the image and text representations for all time steps: $\{\bm{x}^g_{t}, \bm{X}_{t}, \bm{r}^g_{t}, \bm{R}_{t}\}_{t=1}^{|\bm{S}|}$. By viewing image-text pairs in all sequences as independent, Med-ST performs global alignment between matched global image and text features $\{\bm{x}^g_{t}, \bm{r}^g_{t}\}$, and introduces modality-weighted local alignment between matched sequences of patch-wise and token-wise features $\{\bm{X}_{t}, \bm{R}_{t}\}$. For temporal modeling, the global image and text features form two feature sequences, $\bm{X}^g=[\bm{x}^g_{1}, \bm{x}^g_{2}, \cdots, \bm{x}^g_{|\bm{S}|}]$ and $\bm{R}^g=[\bm{r}^g_{1}, \bm{r}^g_{2}, \cdots, \bm{r}^g_{|\bm{S}|}]$, respectively, and Med-ST imposes the cross-modality cycle consistency constraints to align $\bm{X}^g$ and $\bm{R}^g$.




\subsection{Spatial Modeling}
\label{sec:spatialmodeling}


\subsubsection{Image Feature Extraction}
\begin{figure}[t]
\centering
\includegraphics[width=1.0\linewidth]{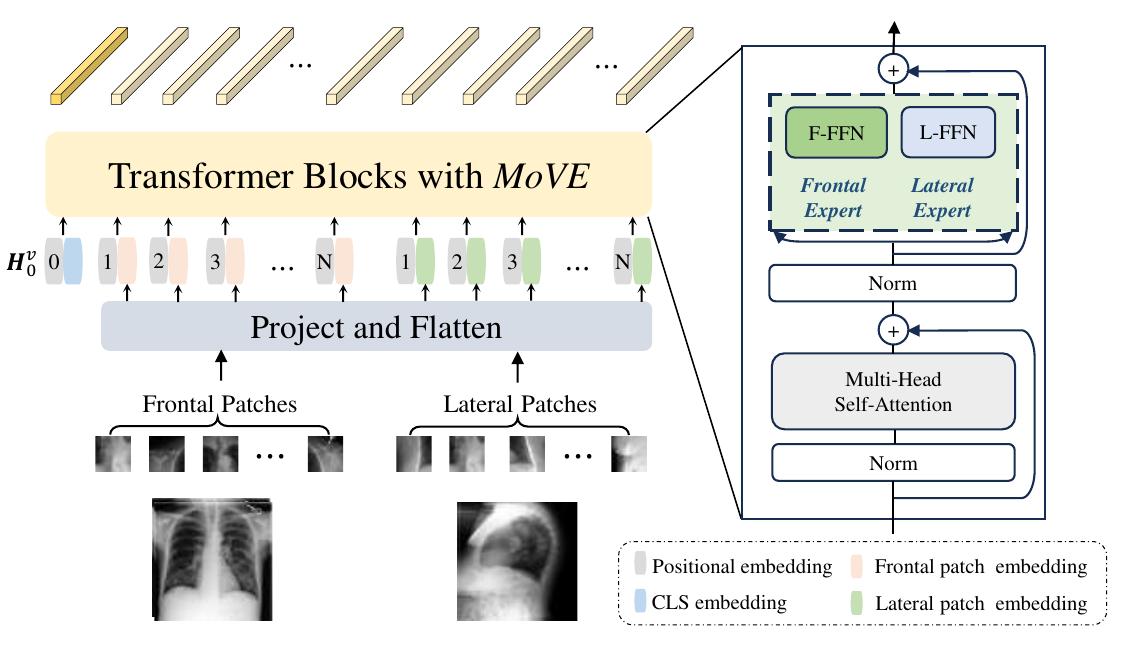}
\caption{The multi-view image encoder. We input both the frontal and lateral views into \emph{MoVE} blocks. The right side shows the structure of \emph{MoVE}.}
\label{fig:spatial}
\end{figure}

Based on Vision Transformer (ViT)~\cite{dosovitskiy2020image}, we develop  a \emph{Mixture of View Experts (MoVE)} architecture as \(f_I\) to encode the multi-view images. As shown in Fig.~\ref{fig:spatial},
for a pair of frontal and lateral images $(\bm{I}^f, \bm{I}^l)$ in a sequence sample $\bm S$, we initially divide them into patches, respectively, which are then flattened through a projector to obtain a series of patch embeddings $\{\bm{p}_j^f\}_{j=1}^{n_p}$, $\{\bm{p}_j^l\}_{j=1}^{n_p}$ with dimension $D$. We add a learnable embedding $\bm{p}_{\text {CLS}}$. To preserve the positional information of patches in each image, we introduce position encoding $\bm{E}_{pos} \in \mathbb{R} ^{({n_p}+1) \times D}$  to both frontal and lateral patches, respectively. All patch embeddings from both views are concatenated to form the embedding sequence $\bm{H}_0^v$ of the image encoder as follows, which serves as the input of the image encoder. 
\begin{equation}
\begin{split}
    \bm{H}_0^v = &\left(\bm{p}_{\text {CLS}} + \bm{E}_{pos}[0,:] \right) \oplus  (\{\bm{p}_j^f\}_{j=1}^{n_p} + \bm{E}_{pos}[1:,:])\\
    &\oplus (\{\bm{p}_j^l\}_{j=1}^{n_p} + \bm{E}_{pos}[1:,:])
\end{split}
\end{equation}
where $\oplus$ is the concatenation operation and $\bm{H}_0^v \in \mathbb{R}^{(2{n_p}+1)\times D}$.

To enable the model to perceive distinctions between the two views, we employ our \emph{MoVE} architecture to replace the standard transformer's block in ViT. Specifically, after obtaining the output  $\bm{H}_{l-1}$ from the previous block, multi-head self-attention layers are performed to capture shared spatial dependencies among all patch embeddings. Then the first ${n_p}+1$ vectors are processed through a Frontal Feedforward Network (F-FFN), while the latter $n_p$ vectors go through a Lateral Feedforward Network (L-FFN). These two FFNs act as distinct experts for capturing complementary information from different perspectives. The outputs of the two experts are concatenated to form the output $\bm{H}_{l}$ of this block. The image encoder consists of $L$ \emph{MoVE}-based blocks. For the output $\bm{H}_{L}$ of the final block, $\bm{x}^g = \bm{H}_{L}[0,:]$ is the global multi-view image feature and $\bm{X}=\bm{H}_{L}[0,:]$ is the sequence of patch-wise features. %




\subsubsection{Cross Modal alignment}
\noindent \textbf{Global Alignment.}
For the $t$-th pair $(\bm{I}^f_{i,t}, \bm{I}^l_{i,t}, \bm{T}_{i,t})$ of the $i$-th sequence in the batch, we obtain $[\bm{x}^g_{i,t}, \bm{X}_{i,t}]=f_I([\bm{I}^f_{i,t}, \bm{I}^l_{i,t}])$. We also use the text encoder to extract the global text feature $\bm{r}^g_{i,t}$ and the sequence of token features $\bm{R}_{i,t}$. Following CLIP~\cite{radford2021learning}, we employ contrastive learning to bring global features of matching images and radiological reports closer while pushing global features of non-matching pairs apart. 
\begin{align}
\label{eq:global}
    l_{i,t}^{v2t} &= - \log \frac{\exp (sim( \bm x_{i,t}^g, \bm r_{i,t}^g)/\tau_1)}{\sum_{b=1}^{B}\sum_{k=1}^{|\bm S_k|} \exp(sim( \bm x_{i,t}^g, \bm r_{b,k}^g)/\tau_1)},  \\
    l_{i,t}^{t2v} &= - \log \frac{\exp (sim( \bm r_{i,t}^g, \bm x_{i,t}^g)/\tau_1)}{\sum_{b=1}^{B}\sum_{k=1}^{|\bm S_k|} \exp(sim(\bm r_{i,t}^g,  \bm x_{b,k}^g)/\tau_1) }.
\end{align}

Here, $B$ is the batch size, $sim(\cdot,\cdot)$ is the similarity function and is measured by dot product. The temperature variable $\tau_1$ is used to scale the logits. So the  objective of Global Alignment is the average of the two losses:
\begin{equation}
    \mathcal{L}_{global} = \frac{1}{2|\mathcal{D}|}\sum_{i=1}^{|\mathcal{D}|}\frac{1}{|\bm S_i|}\sum_{t=1}^{|\bm S_i|} (l_{i,t}^{v2t} + l_{i,t}^{t2v}).
\end{equation}
\label{sec:spatial}

\noindent \textbf{Modality-Weighted Local Alignment.}
Considering that pathological regions may only occupy specific portions of both the image and text, we introduce a novel modality-weighted local alignment between $\bm{X}_{i,t}$ and $\bm{R}_{i,t}$, i.e., paired sequences of patch and token features, to perform fine-grained spatial modeling.

For the $j$-th text token feature \(\bm{r}_{i,t,j}\) in $\bm{R}_{i,t}$, we compute the cosine similarity between \(\bm{r}_{i,t,j}\) and all image patch features in $\bm{X}_{i,t}$, then we apply softmax function to the similarity scores. Thus we get $s = \text{softmax}(\bm{r}_{i,t,j}\bm{X}_{i^,t}^T)$. These scores are then multiplied by the corresponding image patch features and aggregated. As a result, we generate a new textual-attended visual representation $\bm k_{i,t,j}$. Our goal of local alignment is to align \(\bm{r}_{i,t,j}\) and its corresponding textual-attended visual representation $\bm k_{i,t,j}$. 
\begin{equation}
    l_{i,t,j}^{v2t} = - \log \frac{\exp (sim(\bm k_{i,t,j},\bm{r}_{i,t,j})/\tau_2)}{\sum_{c=1}^{W} \exp(sim(\bm k_{i,t,j},\bm{r}_{i,t,c})/\tau_2)},
\end{equation}
\begin{equation}
    l_{i,t,j}^{t2v} = - \log \frac{\exp (sim(\bm {r}_{i,t,j},\bm k_{i,t,j})/\tau_2)}{\sum_{c=1}^{W} \exp(sim(\bm{r}_{i,t,j},\bm k_{i,t,c}/\tau_2)}.
\end{equation}

Acknowledging that different local pairs contain varying amounts of information, those representing pathological semantics should be assigned greater importance. Different from \citet{wang2022multi} using averaged last-layer attention weight in a single modality, we considered the information contained in both visual and textual modalities. For the pair $(\bm k_{i,t,j},\bm{r}_{i,t,j})$, the weights are determined by comparing the value of this pair against the average value of all local pairs in $\bm{X}_{i,t},\bm{R}_{i,t}$. Firstly, we perform an element-wise multiplication of $\bm{r}_{i,t,j}$ and $\bm{k}_{i,t,j}$, i.e., $ \bm o_{i,t,j} =\bm{r}_{i,t,j} \odot \bm{k}_{i,t,j} $. These vectors capture relationships and importance across different dimensions and we form them into a matrix $\bm{\mathcal{O}}_{i,t}=[\bm o_{i,t,1}, \bm o_{i,t,2}, \dots, \bm o_{i,t,W}]$. Then we calculate the mean: $\bar o_{i,t} = \frac{1}{W}\sum_{j=1}^{W} \bm o_{i,t,j}$, from which we derive the weight through softmax attention score.   
\begin{equation}
    w_{i,t,j} = \left(\text{softmax}\left(\frac{\bar o_{i,t}\bm{Q}{\left(\bm{\mathcal{O}}_{i,t}\bm{K} \right)}^{T}}{\sqrt{d}} \right ) \right )_j,
\end{equation}
where $\bm{Q}\in \mathbb{R}^{d\times d}, \bm{K} \in \mathbb{R}^{d \times d}$ are learnable matrices. 
We apply $w_{i,t,j}$ as the weight of $l_{i,t,j}$ and derive the Visual modality-weighted Local Alignment loss $\mathcal{L}_{\text{VLA}}$ as:
\begin{equation}
    \mathcal{L}_{\text{VLA}} = \frac{1}{|\mathcal{D}|W} \sum_{i=1}^{|\mathcal{D}|} \frac{1}{|\bm S_t|} \sum_{t=1}^{|\bm S_t|}  \sum_{j=1}^W w_{i,t,j}(l_{i,t,j}^{v2t} + l_{i,t,j}^{t2v}).
\end{equation}
Correspondingly, we can derive the Texual Modality-Weighted Local Alignment Loss $\mathcal{L}_{\text{TLA}}$. The objective of Local Alignment is:
\begin{equation}
    \mathcal{L}_{local} = \mathcal{L}_{\text{VLA}}+\mathcal{L}_{\text{TLA}}.
\end{equation}

\subsection{Temporal Modeling}
\label{sec:temporalmodeling}
We have observed that there is inherent temporal information in the dataset, which can provide additional supervision signals without the need for extra manual labeling. The objective of our temporal modeling is to use this information to enhance the alignment of image and text sequences that convey the same temporal semantics. 
As mentioned in Sec.~\ref{sec:overview}, for sequence $\bm S$, the global image and text features form two feature sequences, $\bm{X}^g=[\bm{x}^g_{1}, \bm{x}^g_{2}, \cdots, \bm{x}^g_{|\bm{S}|}]$ and $\bm{R}^g=[\bm{r}^g_{1}, \bm{r}^g_{2}, \cdots, \bm{r}^g_{|\bm{S}|}]$, respectively, and Med-ST imposes the cross-modality bidirectional cycle consistency constraints to align $\bm{X}^g$ and $\bm{R}^g$.

\noindent \textbf{Cycle Consistency.} As illustrated in Figure~\ref{fig:framework}, we want to establish bidirectional perception in the sequence. Given an image point $\bm x_t^g$ from $\bm{X}^g$, we first map it to a textual point using function $F$. Subsequently, the result of this mapping is transformed back to an image point through function $G$, thus we get $\bm x_{t'}^g = G(F(\bm x_t^g))$. If $t=t'$, the point $\bm x_t^g$ is cycle-consistent. The bidirectional matching mechanism of cycle consistency can perceive fine-grained contextual differences, thereby establishing a  cross-modal match. Since our model has never explicitly perceived temporal sequence information, directly learning such information is challenging. Therefore, we adopt a novel bidirectional learning approach from simple to complex. 

\noindent \textbf{Forward Mapping Classification (FMC).}
We first calculate the distance between $\bm x_t^g$ and all points in $\bm R^g$ and then we get the soft nearest neighbor $F(\bm x_t^g)$ of \( \bm x_t^g \):
\begin{equation}
\label{eq:fx}
    F(\bm x_t^g) = \sum_{z=1}^{|\bm S|}\alpha_z\bm r_z^g,  ~\alpha_z = \frac{\exp(-\langle \bm x_t^g, \bm r_z^g \rangle)}{\sum_{k=1}^{|\bm S|}\exp(- \langle \bm x_t^g, \bm r_k^g \rangle)}.
\end{equation}
Since our sequence is paired, unlike~\citet{dwibedi2019temporal}, 
we can introduce constraints to the forward process. As  our current model lacks temporal awareness, we incorporate a relatively simple classification loss in the forward process. We treat all features in sequence $\bm R^g $ as different classes and classify $\bm x_t^g$ accordingly. So the predicted output $\hat y_t = \alpha_t$. The ground truth $\bm{y}$ is represented by a one-hot vector, in which the $t$-th position is set to 1.
\begin{equation}
    \mathcal{L}^{cls}_{t} = -\sum_{z=1}^{|\bm S|}y_z \log(\hat{y_z}).
\end{equation}
As for sequence $\bm S$, the objective of FMC is as follows:
\begin{equation}
    \mathcal{L}_{\text{FMC}}^I = \sum_{t}^{|\bm S|} \mathcal{L}^{cls}_{t}.
\end{equation}


\noindent \textbf{Reverse Mapping Regression (RMR).}
Since FMC enables cycle consistency from a relatively easy classification perspective, we want to incorporate a more complex objective in the process of reverse mapping. We impose greater penalties on predictions that are temporally distant according to the distance between $F(\bm x_t^g)$ and $\bm x_t^g$. To address this, we propose RMR. Initially, we obtain the distribution of similarities $\{\beta_z\}$ between $F(\bm x_t^g)$ and  all points in $\bm X^g$. 
\begin{equation}
 \beta_z = \frac{\exp{(-\langle F(\bm x_t^g), \bm x_z^g \rangle )}}{\sum_{k=1}^{|\bm S|} \exp {(-\langle F(\bm x_t^g) , \bm x_k^g \rangle})}. 
\end{equation}
Representations that are temporally closer tend to be more similar. Thus the further the distance from timestep $t$, the smaller the similarity should be. 
So we impose a Gaussian prior on $\beta$.
Our goal is to maximize $\beta_t$ and ensure that the similarity distribution peaks at $t$. 
Therefore, we optimize $\frac{|t-\mu|^2}{\sigma^2}$. A regularization term $\log(\sigma)$ is added to ensure the peak of the distribution remains intact and avoid a reduction in variance resulting from this optimization. Since the temporal information contained in the sequence may have a large span, for cases with larger errors, we optimize $|t - \mu|$. So the objective of RMR for sequence $\bm S$ is as follows:
\begin{equation}
\label{eq:rmr}
\mathcal{L}_{\text{RMR}}^I = \begin{cases} 
  \displaystyle \frac{|t - \mu|^2}{\sigma^2} + \lambda \log(\sigma) & , |t-\mu| \leq \delta, \\
  \delta|t-\mu|-\displaystyle\frac{\delta^2}{2} & , \text{otherwise,}
\end{cases} 
\end{equation}
where \( \mu = \sum_{z=1}^{|\bm S|} \beta_z \cdot z \) and \( \sigma^2 = \sum_{z=1}^{|\bm S|} \beta_z \cdot (z - \mu)^2 \),  \( \lambda \) denotes the regularization weight, and \(\delta\) is a hyperparameter.

Therefore, for all sequences in $\mathcal{D}$, we sum these two objectives to obtain $\mathcal{L}_{temporal}^I = \mathcal{L}_{\text{RMR}}^I + \mathcal{L}_{\text{FMC}}^I$.
Similarly, starting from a point in the text sequence, we can get $\mathcal{L}_{temporal}^T$. So the objective of temporal modeling is:
\begin{equation}
    \mathcal{L}_{temporal} = \mathcal{L}_{temporal}^I + \mathcal{L}_{temporal}^T. 
\end{equation}


\subsection{Overall Objective}
Therefore, our overall objective is denoted as $\mathcal{L}$ in Eq.~\ref{eq:overall}. By optimizing this objective, our model can perceive  spatial and temporal information with greater granularity.
\begin{equation}
\label{eq:overall}
    \mathcal{L} = \mathcal{L}_{global} + \lambda_1 \mathcal{L}_{local} + \lambda_2 \mathcal{L}_{temporal},
\end{equation} 
where $\lambda_1$ and $\lambda_2$ are hyperparameters that control the weights of the loss functions.

\section{Experiments}
\label{sec:experiment}



\begin{table*}[t]
\caption{Temporal image classification results on the MS-CXR-T benchmark across five findings under 10-fold cross-validation setting.  We run  with different seeds three times and calculate the mean and standard deviations. \textbf{Avg. Acc.} stands for average accuracy. PL.effusion denotes pleural effusion. We report the accuracy [\%]. Best results are in boldface.}
\label{tab:10fold}
\vskip -0.1in
\label{sample-table}
\begin{center}
\begin{small}
\begin{tabular}{l|ccccc|c}
\toprule

Model           & Consolidation & Edema & Pl.effusion & Pneumonia & Pneumothorax & \textbf{Avg. Acc.}    \\ \midrule
ConVIRT-MIMIC~\cite{zhang2022contrastive}   & 46.62\tiny{$\pm$ 1.03}     & 57.04\tiny{$\pm$ 1.00}  & 54.50\tiny{$\pm$ 0.19}              & 61.09\tiny{$\pm$ 1.54}      & 48.49\tiny{$\pm$ 0.22}         & 53.55\tiny{$\pm$ 0.36}   \\
GLoRIA-CheXpert~\cite{huang2021gloria} & 49.13\tiny{$\pm$ 1.52}          & 53.67\tiny{$\pm$ 0.61}  & 49.56\tiny{$\pm$ 1.84}             & 58.11\tiny{$\pm$ 2.21}    & 45.64\tiny{$\pm$ 2.49}        & 51.22\tiny{$\pm$ 1.35}  \\
GLoRIA-MIMIC~\cite{huang2021gloria}    & 44.99\tiny{$\pm$ 0.47}         & 49.13\tiny{$\pm$ 1.54} & 47.85\tiny{$\pm$ 3.08}             & 60.18\tiny{$\pm$ 1.11}     & 41.53\tiny{$\pm$ 0.92}        & 48.74\tiny{$\pm$ 0.84}  \\
MGCA~\cite{wang2022multi}            & 50.79\tiny{$\pm$ 0.44}         & 62.71\tiny{$\pm$ 0.24} & 57.17\tiny{$\pm$ 0.69}             & 63.73\tiny{$\pm$ 0.89}     & 52.16\tiny{$\pm$ 1.02}       & 57.31\tiny{$\pm$ 0.34} \\

MedCLIP~\cite{wang2022medclip}         & 50.32\tiny{$\pm$ 0.65}         & 54.53\tiny{$\pm$ 2.82} & 56.61\tiny{$\pm$ 0.65}             & 58.94\tiny{$\pm$ 2.62}    & 45.19\tiny{$\pm$ 1.47}        & 53.12\tiny{$\pm$ 0.13}  \\

MRM~\cite{wang2022medclip}         & 46.33\tiny{$\pm$ 2.89}         & 49.09\tiny{$\pm$ 5.70} & 49.13\tiny{$\pm$ 0.69}             & 55.14\tiny{$\pm$ 1.42}    & 45.48\tiny{$\pm$ 2.97}        & 49.03\tiny{$\pm$ 0.80} \\

BioViL~\cite{boecking2022making}          & 56.40\tiny{$\pm$ 0.24}  & 57.26\tiny{$\pm$ 0.77}  & 54.51\tiny{$\pm$ 0.39} & 67.10\tiny{$\pm$ 0.01}      & 55.30\tiny{$\pm$ 0.21}         & 58.11\tiny{$\pm$ 0.02}   \\

\hline
\multicolumn{7}{l}{\emph{\small{Temporal-based}}} \\
BioViL-T~\cite{bannur2023learning}        & 56.93\tiny{$\pm$ 1.77}         & 61.55\tiny{$\pm$ 0.95} & 53.94\tiny{$\pm$ 0.89}             & \textbf{67.24}\tiny{$\pm$ 0.20}     & \textbf{55.46}\tiny{$\pm$ 0.01}        & 59.02\tiny{$\pm$ 0.34}  \\
\rowcolor{gray!20}Med-ST           & \textbf{60.57}\tiny{$\pm$ 1.18}       &\textbf{67.35}\tiny{$\pm$ 0.32} & \textbf{58.47}\tiny{$\pm$ 1.50}         & {65.00\tiny{$\pm$ 0.34}}     & {54.18}\tiny{$\pm$ 0.81}     & \textbf{61.12}\tiny{$\pm$ 0.34}  \\

\bottomrule
\end{tabular}
\end{small}
\end{center}
\vskip -0.1in
\end{table*}

\begin{table}[t]
\caption{Temporal sentence similarity classification results  on RadGraph subset of MS-CXR-T sentence similarity benchmark. Accuracy and AUROC scores are reported. Best results are in boldface.}
\label{tab:sentence}
\begin{center}
\setlength{\tabcolsep}{12pt}
\vskip -0.1in
\begin{tabular}{l|c|cc}
\toprule
Model    & MLM &Acc. & AUROC \\ \midrule
MedCLIP &   $\times$ &66.41         & 52.66 \\
MGCA     &  $\times$  &75.42    & 76.38   \\
BioViL   &  \checkmark  &69.49    & 68.92   \\
BioViL-T &  \checkmark  &78.81    & 81.39   \\
\rowcolor{gray!20}Med-ST   &  $\times$   & \textbf{83.76}    & \textbf{84.60}   \\ \bottomrule
\end{tabular}
\end{center}
\vskip -0.1in

\end{table}

\begin{table}[]
\caption{Zero-shot classification results on RSNA. We report Accuracy, F1 and AUROC scores. Best results are  highlighted in bold.}
\label{tab:zeroshot}
\vspace{-8pt}
\begin{center}
\setlength{\tabcolsep}{12pt}
\begin{tabular}{l|ccc}
\toprule
Model        & Acc.   & F1    & AUROC \\ \midrule
MGCA     & 67.25 & 54.87 & 73.97 \\
BioViL   & 64.10 & 55.19  & 75.27  \\
BioViL-T & 63.23 & 54.90 & 75.12 \\
\rowcolor{gray!15}Med-ST   & \textbf{68.37} & \textbf{57.63} & \textbf{77.14} \\ 
\bottomrule
\end{tabular}
\end{center}
\vskip -0.15in

\end{table}

\begin{table}[]
\caption{Medical image classification results on COVIDx datasets with 1\%, 10\% and 100\% training data. }
\label{tab:covidx}
\vspace{-8pt}
\begin{center}
\setlength{\tabcolsep}{12pt}
\begin{tabular}{l|ccc}
\toprule
Method       & 1\%  & 10\% & 100\% \\ \midrule
GLoRIA       & 67.3 & 77.8 & 89.0  \\
ConVIRT      & 72.5 & 82.5 & 92.0  \\
GLoRIA-MIMIC & 66.5 & 80.5 & 88.8  \\
MedKLIP      & 74.5 & 85.2 & 90.3  \\
MGCA         & \textbf{74.8} & 84.8 & 92.3  \\
\rowcolor{gray!15}Med-ST       & 71.2 & \textbf{87.7} & \textbf{93.0}  \\ \bottomrule
\end{tabular}
\end{center}
\vskip -0.1in

\end{table}

\begin{table*}[]
\caption{Ablation study on objectives and lateral view use.}
\vspace{-8pt}
\label{tab:albloss}
\vskip -10pt
\begin{center}
\scalebox{0.93}{
\begin{tabular}{lcc|c|ccccc|c}
\toprule
$\mathcal{L}_{global}$ & $\mathcal{L}_{temporal}$ & $\mathcal{L}_{local}$ & Lateral  & Consolidation & Edema & Pl.effusion & Pneumonia & Pneumothorax & Avg. Acc. \\ \midrule
\checkmark &  &\checkmark &\checkmark  &55.75\tiny{$\pm$ 2.52}  &62.20\tiny{$\pm$ 1.00} & 57.91\tiny{$\pm$ 1.20} & 63.88\tiny{$\pm$ 0.79} & 54.53\tiny{$\pm$ 0.99} & 58.85\tiny{$\pm$ 0.61} \\
\checkmark & \checkmark & &\checkmark  &54.89\tiny{$\pm$ 0.51} &65.30\tiny{$\pm$ 0.91}  &57.42\tiny{$\pm$ 0.39}  &66.40\tiny{$\pm$ 0.19}  &53.40\tiny{$\pm$ 0.96}   &59.49\tiny{$\pm$ 0.30}  \\
\checkmark & \checkmark &\checkmark& &55.09\tiny{$\pm$ 1.34} &63.32\tiny{$\pm$ 0.78}  &58.06\tiny{$\pm$ 1.40} &53.58\tiny{$\pm$ 0.40} &50.25\tiny{$\pm$ 0.99} &58.06\tiny{$\pm$ 0.29}  \\
\rowcolor{gray!15}\checkmark & \checkmark &\checkmark & \checkmark &60.57\tiny{$\pm$ 1.18} & 67.35\tiny{$\pm$ 0.32} &58.47\tiny{$\pm$ 1.50}& 65.00\tiny{$\pm$ 0.34}&54.18\tiny{$\pm$ 0.81}&61.12\tiny{$\pm$ 0.34} \\
\bottomrule
\end{tabular}
}
\end{center}
\vskip -0.1in
\end{table*}

\subsection{Pre-training Setup}
\noindent \textbf{Dataset.} We pretrain our Med-ST framework on \textbf{MIMIC-CXR} dataset~\cite{johnson2019mimic}. 
It is a publicly available dataset that includes paired chest X-ray images and radiology reports. Sometimes images incorporate both frontal and lateral views. All pairs have attributes of \texttt{studydate} and \texttt{studytime}, which allow us to construct sequences. We pre-process the dataset following~\citet{wang2022multi}, including image transformation and text tokenization, resulting in 232$k$ image-text pairs. We preserve lateral views and thus $114k$ pairs include lateral images.

\noindent \textbf{Implementation Details.}
Our code is implemented using PyTorch~\cite{paszke2019pytorch}. 
The pre-training is performed using two GeForce RTX 3090 GPUs.
Due to the effectiveness of pre-trained language models, we utilized BioClinicalBERT~\cite{alsentzer2019publicly} as the text encoder. For unified modal architecture design, we defaulted to using ViT-B/16~\cite{dosovitskiy2020image} as the image encoder backbone. We use the AdamW optimizer~\cite{loshchilov2017decoupled} with a learning rate of 4e-5 and weight decay of 0.05. We employed a linear warmup with cosine annealing scheduler~\cite{loshchilov2016sgdr}, initializing the learning rate to 1e-8 and the warmup epoch to 20.  The $\tau_1$ is 0.07 and $\tau_2$ is 0.1.  We set $\lambda_1$, $\lambda_2$  to 1, 1 in Eq.~\ref{eq:overall} and $\delta$=2, $\lambda$=0.001 in Eq.~\ref{eq:rmr}. For more details, please refer to Appendix~\ref{app:pretrain}.

\subsection{Experimental Setup}
\subsubsection{Dataset}
\noindent \textbf{MS-CXR-T benchmark}~\cite{bannur2023learning} is a multi-modal benchmark dataset aimed at evaluating biomedical vision-language processing models in radiology, specifically through two temporal tasks: image classification of chest X-rays and analysis of sentence similarity.

\noindent \textbf{RSNA dataset}~\cite{shih2019augmenting}  is a publicly available medical imaging dataset. It contains chest radiographs with annotations for the presence or absence of pneumonia.

\noindent \textbf{COVIDx}~\cite{wang2020covid} is a dataset with over 30,000 chest X-ray (CXR) images from over 16,600 patients, including 16,490 positive COVID-19 cases. It's used for a three-class classification task: categorizing radiographs as COVID-19, non-COVID pneumonia, or normal.

\subsubsection{Downstream Tasks}
\noindent \textbf{Temporal Image Classification.}
It includes 1,326 labeled chest X-rays across five findings (Consolidation, Edema, Pleural Effusion, Pneumonia, and Pneumothorax), categorized into three stages of disease progression: \texttt{Improving, Stable, Worsening}. Due to the limited data size, we choose an SVM~\cite{boser1992training} classifier and employ cross-validation. We report the accuracy for each findings and the average classification accuracy under the setting of 5-fold and 10-fold.

\noindent \textbf{Temporal Sentence Similarity.}
It includes 361 sentence pairs for evaluating temporal-semantic similarity. The task focuses on binary classification of pairs as paraphrases or contradictions regarding disease progression. This dataset includes two subsets: RadGraph and Swaps.We report our results on the RadGraph subset since it is more challenging. For more experimental settings and results on the Swaps subset, please refer to the Appendix~\ref{app:sentence}.

\noindent \textbf{Zero-shot Image Classification.}
We test our zero-shot classification results on RSNA. We use the dataset division from previous work~\cite{wang2022multi,huang2021gloria} and also employ the prompt construction of BioViL-T. We report our classification results on the test set.

\noindent \textbf{Medical Image Classification.}
Our model is tested on the COVIDx dataset to evaluate its efficacy in medical image classification. As with prior methods~\cite{wang2022multi}, we fine-tune a classification head using different proportions of training data and then evaluate the classification performance on the test set.

\subsection{Results}
\noindent \textbf{Temporal Image Classification Results.}
In Table~\ref{tab:10fold}, we report our performance on the task of temporal image classification under a 10-fold cross-validation setup. Our method achieves the highest average  accuracy, surpassing BioViL-T~\cite{bannur2023learning}, which also utilizes temporal information, by 2.10\%. Across five different findings, our method achieves the highest scores in three categories, particularly in the edema category, where our approach exceedes BioViL-T by 5.80\%. Table~\ref{tab:5fold} also validates our performance. We also achieve the best average accuracy  in the experimental setup with a fold number of 5 in Table~\ref{tab:5fold}. Our method consistently achieves the best results in different cross-validation setups. This demonstrates that through our temporal modeling, the model is able to capture the semantics of temporal changes and integrate this information into the model, making it more suitable for temporal tasks.

\noindent \textbf{Temporal Sentence Similarity Classification Results.}
In Table~\ref{tab:sentence}, we report our performance on the temporal text classification task. Our method achieves the best results in both accuracy and AUROC metrics. It surpasses BioViL-T by 4.95\% in accuracy and by 3.21\% in the AUROC metric. The BioViL-T method not only utilizes temporal information but also performs specialized mask language modeling. Additionally, the RadGraph dataset is a relatively complex text classification task, hence our results in this experiment thoroughly demonstrate the effectiveness of our approach. Through temporal modeling, our method acquires the capability to perceive changes, and with fine-grained local alignment, it more comprehensively understands the semantic context of the text.

\noindent \textbf{Zero-shot Image Classification Results.}
Table~\ref{tab:zeroshot} shows our results of zero-shot medical image classification task on RSNA dataset. Our method also achieves the best performance. This indicates that our performance improvement is not limited to temporal tasks but also benefits static classification tasks. Our model outperforms previous methods by 1\%, 2\%, and 1\% in the accuracy, F1 score, and AUROC metrics, respectively. These consistent improvements across metrics suggest that spatial and temporal modeling can enhance the model's representational capabilities. Since the logits in zero-shot classification are based on the similarity between image features and text prompts, it also indicates that our image and text encoders have learned more fine-grained and consistent representations.

\noindent \textbf{Medical Image Classification Results.}
We present our classification results on COVIDx in Table~\ref{tab:covidx}. Our method achieved commendable results using both 10\% and 100\% of the training data. The outcomes in the classification task also demonstrate that the gains from our approach are evident in static tasks. Because our model absorbs a wider range of supervision signals, it may fail to capture the relevant information with only 1\% of the training data available.

\begin{figure}[!t]
	\begin{center}
 \vskip 8pt
		\includegraphics[width=1.0\linewidth]{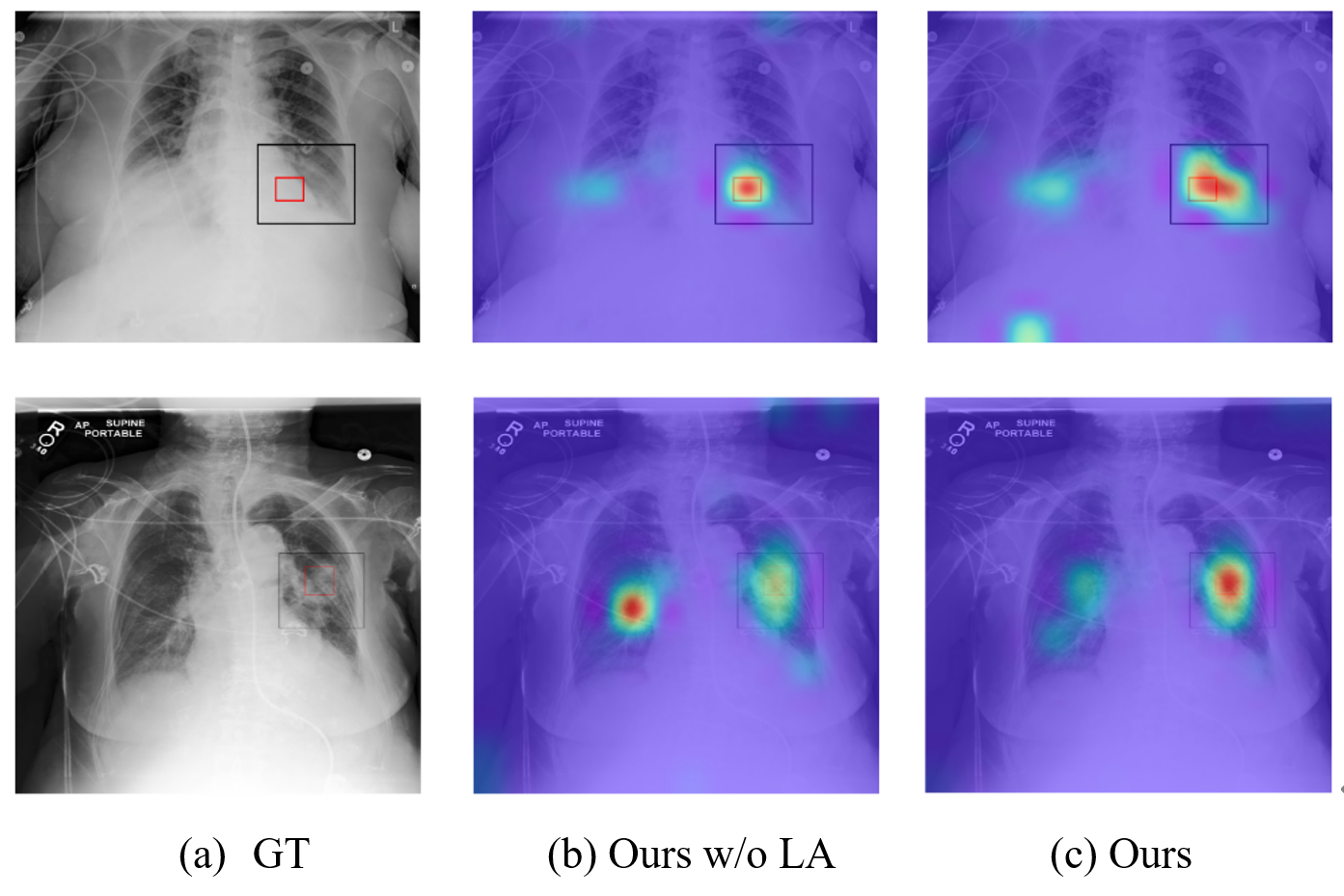}
	\end{center}
	\caption{Attention visualization with reference patch (red box) in the bounding box (black box). LA stands for local alignment. The bounding box represents the main pathological area annotated manually. It can be observed that our method achieves better correspondence with the bounding box after local alignment.}
 \label{fig:visualize}
\end{figure}
\subsection{Analysis of Our Framework}
\noindent \textbf{Ablation Study.}
Table~\ref{tab:albloss} displays the ablation results of our various components on the temporal image classification task. It can be observed that the best results are obtained when all objectives are combined with the use of lateral views. The first two rows indicate the effectiveness of our temporal modeling and local alignment objectives. The comparison between the third and fourth rows highlights the efficacy of using lateral views, and when all objectives are combined, we can achieve further improvements. 

\noindent \textbf{Visualization.} 
Figure~\ref{fig:visualize} presents the attention visualization of our model based on the reference patch (red box). This patch, contained within the bounding box (black box), encapsulates the corresponding textual information. The text corresponding to the first row is ``Left basilar consolidation seen" and the second row corresponds to ``patchy consolidation in the mid left lung". It can be observed that our method, without local alignment, can only perceive certain areas. However, after applying modality-weighted local alignment, our method is able to recognize the pathological areas in the context, which correspond well with the bounding box. This demonstrates that our fine-grained modeling can indeed learn local alignment based on the amount of information contained in local pairs.

\section{Conclusion}
\label{sec:conclusion}
In this study, we present the Med-ST framework, which employs explicit spatial and temporal modeling along with comprehensive strategies for global and local alignment, and cross-modal bidirectional cycle  consistency. This approach not only achieves superior performance in tasks including temporal sequences but also enhances the representation learning for medical image classification tasks. Med-ST emerges as a promising approach for medical multimodal pretraining, capitalizing on the extensive information present in multimodal datasets to acquire nuanced spatial and temporal insights.

\section*{Acknowledgements}
This work was supported in part by the National Natural Science Foundation of China No. 62376277, No. 61976206 and No. 62222215, and Beijing Outstanding Young Scientist Program NO. BJJWZYJH012019100020098.

\section*{Impact Statement} Although the dataset we used contains temporal information, the specific dates are anonymized, eliminating the risk of privacy infringement. The Med-ST framework introduced in our paper models from both spatial and temporal perspectives, effectively simulating the behavior of clinicians in practice. We offer a promising approach that not only diagnoses abnormalities in chest radiographs but also analyzes historical data to make decisions. This can help reduce the workload on doctors and advance the development of medical diagnostics, making diagnosis more intelligent and accessible, especially in underdeveloped areas.

\bibliography{example_paper}
\bibliographystyle{icml2024}

\newpage
\appendix
\onecolumn

\section{Pre-training Details}
\label{app:pretrain}
We sort all cases of the same patient according to the studyDate attribute in chronological order. Then we divide them into sequences of length 4, treating any remaining part that is less than 4 as a sequence of length 1. In other words, our dataset contains sequences of either four image-text pairs or a single image-text pair. During training, a batch randomly includes sequences of length 4 or 1. Global alignment and local alignment use all image-text pairs in all sequences, while sequence modeling employs all sequences of length 4.

Table~\ref{tab:hyperparameters} outlines the hyperparameters used during pre-training. We opt for MGCA weights initialization, leveraging their high performance to ensure faster convergence, particularly aiming for time efficiency. We use PyTorch Lightning and  an early stopping mechanism to optimize training duration.
\begin{table}[h]
\caption{Hyperparameters of Pre-training}
\vskip -0.1in
\label{tab:hyperparameters}
\begin{center}
\begin{tabular}{ll}
\toprule
\multicolumn{2}{l}{Hyperparameters}       \\ \midrule
Pre-training epochs     & 10              \\
Batch size per GPU      & 10              \\
Number of GPUs          & 2               \\
Learning Rate           & 4e-5            \\
Learning Rate Optimizer & CosineAnnealing \\
Weight Decay            & 0.05            \\ \bottomrule
\end{tabular}
\end{center}

\end{table}

\section{Downstream Tasks}
\subsection{Temporal Image classification Similarity}
\subsubsection{Experimental Details}
We split the dataset according to the annotated disease types, i.e., Consolidation, Edema, Pleural Effusion, Pneumonia, and Pneumothorax. And then, for each subtask, we first obtain the features of two images, concatenate them, and use an SVM for classification, employing cross-validation to get the accuracy for each category. The task is to  categorize the concatenated feature into three stages of disease progression: \texttt{Improving, Stable, Worsening}. Subsequently, we calculate the average accuracy across five types of findings. We use publicly available pre-trained models from others and determine their accuracy on this task in the same manner.

\begin{table*}[t]
\caption{Temporal image classification results on the MS-CXR-T benchmark under 5-fold cross-validation setting. MLM stands for masked language modeling objective. Best results are highlighted in bold.}
\label{tab:5fold}
\vskip -0.3in
\begin{center}
\begin{tabular}{l|ccccc|c}
\toprule

Model           & Consolidation & Edema & Pl.effusion & Pneumonia & Pneumothorax & \textbf{Avg. Acc.}    \\ \midrule
ConVIRT-MIMIC & 46.97\tiny{$\pm$ 1.68}     & 56.03\tiny{$\pm$ 1.07}  & 54.98\tiny{$\pm$ 2.75}              & 60.62\tiny{$\pm$ 2.51}      & 46.61\tiny{$\pm$ 2.51}         & 53.04\tiny{$\pm$ 0.62}   \\
GLoRIA-CheXpert& 50.10\tiny{$\pm$ 1.94}          & 51.63\tiny{$\pm$ 0.46}  & 51.09\tiny{$\pm$ 0.72}       & 60.49\tiny{$\pm$ 1.05}    & 44.10\tiny{$\pm$ 1.18}        & 51.48\tiny{$\pm$ 0.84}  \\
GLoRIA-MIMIC   & 44.32\tiny{$\pm$ 2.51}         & 49.64\tiny{$\pm$ 1.74} & 48.74\tiny{$\pm$ 1.69}             & 58.63\tiny{$\pm$ 2.52}     & 42.17\tiny{$\pm$ 1.66}        & 48.70\tiny{$\pm$ 0.86}  \\
MGCA          & 52.43\tiny{$\pm$ 1.82}         & 63.30\tiny{$\pm$ 1.07} & 56.13\tiny{$\pm$ 1.97}             & 64.83\tiny{$\pm$ 0.21}     & 52.78\tiny{$\pm$ 1.37}       & 57.90\tiny{$\pm$ 0.30} \\

MedCLIP        & 49.44\tiny{$\pm$ 2.51}         & 52.27\tiny{$\pm$ 3.98} & 53.79\tiny{$\pm$ 0.59}             & 58.92\tiny{$\pm$ 1.43}    & 41.87\tiny{$\pm$ 2.38}        & 51.25\tiny{$\pm$ 0.65}  \\

MRM        & 50.10\tiny{$\pm$ 1.16}         & 49.37\tiny{$\pm$ 3.84} & 48.97\tiny{$\pm$ 0.61}             & 55.12\tiny{$\pm$ 1.09}    & 48.23\tiny{$\pm$ 0.47}        & 50.36\tiny{$\pm$ 1.01} \\

BioViL        & 57.56\tiny{$\pm$ 1.17}  & 57.27\tiny{$\pm$ 0.72}  & 54.10\tiny{$\pm$ 0.65} & 66.95\tiny{$\pm$ 0.53}      & 54.82\tiny{$\pm$ 0.22}         & 58.14\tiny{$\pm$ 0.41}   \\

\hline
\multicolumn{7}{l}{\emph{\small{Temporal-based}}} \\
BioViL-T       & 56.73\tiny{$\pm$ 1.76}         & 59.52\tiny{$\pm$ 1.09} & 52.80\tiny{$\pm$ 0.90}             & \textbf{66.67}\tiny{$\pm$ 0.60}     & \textbf{55.61}\tiny{$\pm$ 0.23}        & 58.26\tiny{$\pm$ 0.42}  \\

\rowcolor{gray!20}Med-ST           & \textbf{60.72}\tiny{$\pm$ 2.02}       &\textbf{65.56}\tiny{$\pm$ 1.24} & \textbf{57.99}\tiny{$\pm$ 0.30}         & {64.98\tiny{$\pm$ 0.60}}     & {53.71}\tiny{$\pm$ 1.83}     & \textbf{60.59}\tiny{$\pm$ 0.72}  \\

\bottomrule
\end{tabular}
\end{center}
\end{table*}

\begin{table}[t]
\caption{Temporal sentence similarity classification results
on Swaps subset.}
\label{tab:swaps}
\begin{center}
\begin{tabular}{lcc}
\toprule
Model    & Accuracy & ROC-AUC \\ \midrule
MedCLIP &  59.59         & 48.33\\
MGCA     & 68.16    & 71.97   \\
BioViL   & 81.63    & 90.00 \\
BioViL-T & 94.29    & 97.59   \\
Med-ST   & 76.73    & 86.12   \\
\bottomrule
\end{tabular}
\end{center}
\end{table}

\subsection{Temporal Sentence Similarity}
\label{app:sentence}
\subsubsection{Experimental Details}
The task has two subsets. The first subset, generated by RadGraph, identifies paraphrase and contradiction sentence pairs through analyzing graph representations of sentences. The second subset is created using the Swaps method, which involves changing temporal keywords within sentences to produce paraphrases and contradictions. Swaps approach creates pairs that differ subtly in their temporal semantics while pairs obtained through the RadGraph method exhibit greater variation. Table~\ref{app:sentence} are two examples of sentence pairs from each subset.
We divide the dataset into two subsets based on the construction of text pairs into  RadGraph and Swaps and conduct tests on each subset. Then, firstly, we randomize the dataset and then obtain the AUROC results based on the similarity between text pairs. Subsequently, through ten-fold cross-validation, we select the similarity threshold corresponding to the highest accuracy on the validation set. We then test on the entire set to obtain the accuracy. We use publicly available pre-trained models from others and follow the same process to determine their performance on this task. 
\subsubsection{Results}
Tables~\ref{tab:sentence} and Table~\ref{tab:swaps} respectively show the experimental results on the two subsets. Our method achieves better results on the more complex dataset. For the Swaps dataset, methods BioViL  and BioViL-T achieve better results due to the use of masked language modeling.

\begin{table*}[]
\caption{Examples of sentence pairs from the MS-CXR-T temporal sentence similarity benchmark.}
\label{tab:examples}
\begin{tabular}{p{1.5cm}p{2cm}p{6cm}p{6cm}}

\toprule
                          & Label         & Sentence 1                                                                      & Sentence 2                                                                              \\ \midrule
\multirow{2}{*}{Swaps}    & Paraphrase    & ``Nearly resolved bilateral pleural effusions."                                     & ``Nearly cleared bilateral pleural effusions."                                             \\
                          & Contradiction & ``Bibasal atelectasis is unchanged."                                                & ``Bibasal atelectasis is new."                                                             \\ \midrule
\multirow{5}{*}{RadGraph} & Paraphrase    & ``Moderate pulmonary edema is exaggerated by low lung volumes, but also worsened." & ``Lung volumes are lower exaggerating what is at least worsened moderate pulmonary edema." \\
                          & Contradiction & ``Small right pneumothorax has developed at the base of the right lung."            & ``Small pneumothorax at the base of the right lung is unchanged."                          \\ \bottomrule
\end{tabular}
\end{table*}

\subsection{Zero-shot Image Classification}
\subsubsection{Text Prompt}
We use the  same text prompt as BioViL-T for \texttt{pneumonia}.

\begin{minipage}{\linewidth}
\texttt{\textbf{pos\_query} = [
        'Findings consistent with pneumonia',
        'Findings suggesting pneumonia',
        'This opacity can represent pneumonia',
        'Findings are most compatible with pneumonia',
    ],\\
    \textbf{neg\_query} = [
        'There is no pneumonia',
        'No evidence of pneumonia',
        'No evidence of acute pneumonia',
        'No signs of pneumonia',
    ]
}
\end{minipage}

\subsubsection{Experimental Details}
We use the dataset split from previous work~\cite{huang2021gloria,wang2022multi} and test on the test set. The text embedding is the average of four prompt embeddings, thus we obtain two text features representing the presence and absence of pneumonia. After obtaining the image features, we determine the predicted category based on the similarity between the image features and the two text features.



\section{More Visualization}
\subsection{t-SNE visualizations results of image features}
\begin{figure*}[!t]
\begin{center}
\includegraphics[width=1.0\linewidth]{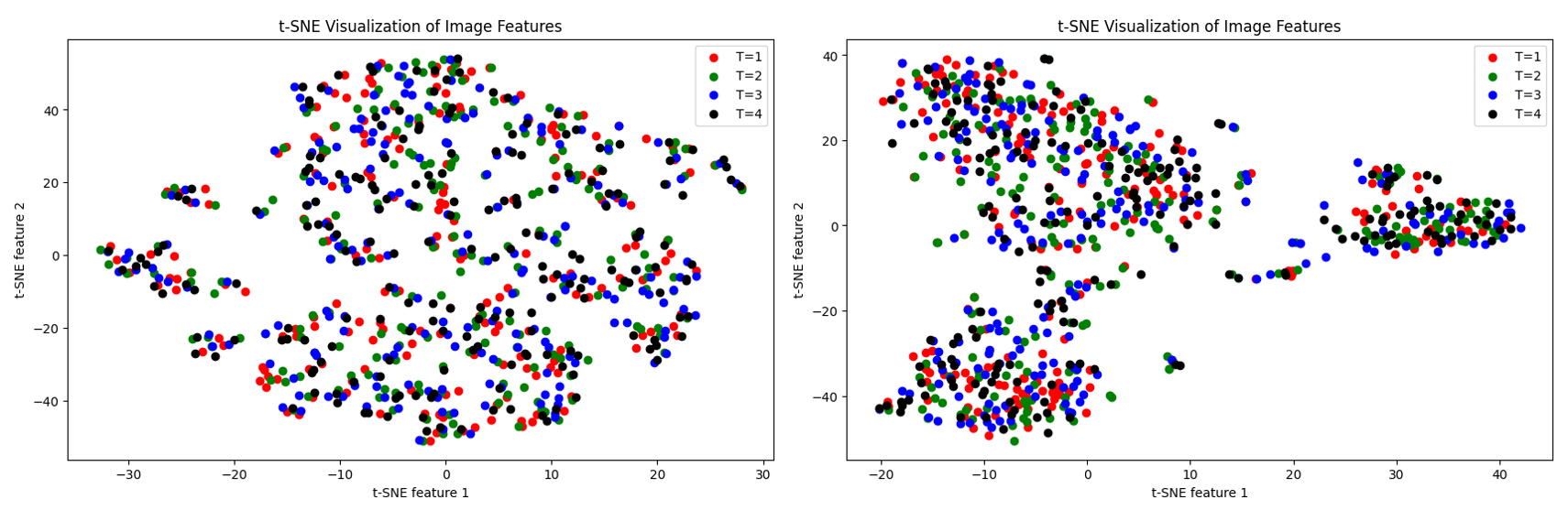}
\end{center}
\vskip -0.1 in
\caption{t-SNE visualizations results of image features at different time steps for 200 sequences. The left side is from MGCA, and the right side is ours.}
\label{fig:tsne}
\end{figure*}

Figure~\ref{fig:tsne} shows the visualization results of image features at different time steps for 200 sequences. The left side is from MGCA, and the right side is ours. Each category represents different time points in the sequence. From the figure, it can be seen that the feature distribution of MGCA is relatively uniform, with no discernible trend of change. In contrast, the trends between the four time points in our diagram are consistent, indicating that our model has learned time-related information.

\subsection{Visualization of the learned distribution of $\beta$}
\begin{figure*}[!t]
\begin{center}
\includegraphics[width=1.0\linewidth]{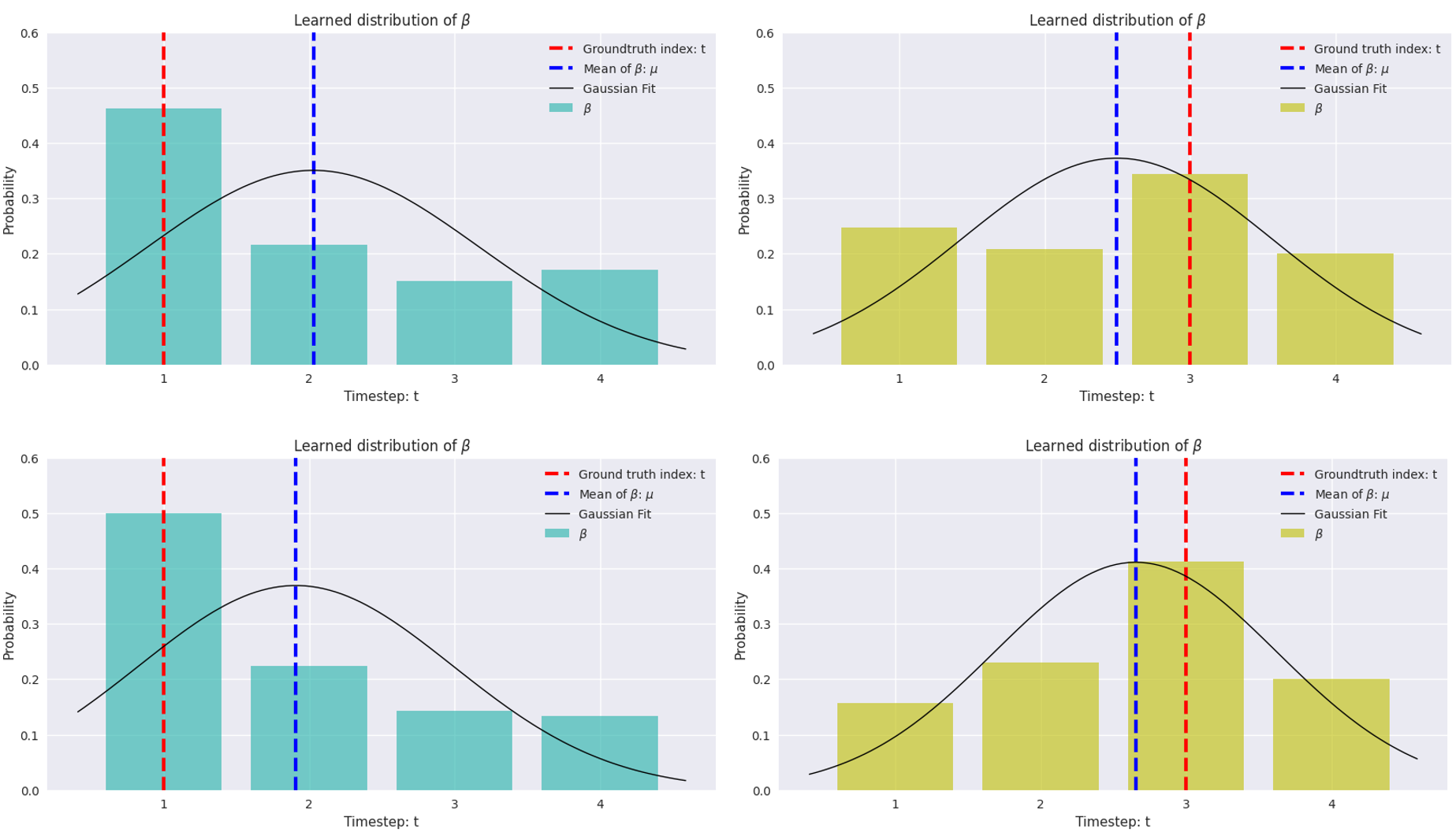}
\end{center}
\vskip -0.1 in
\caption{Visualization of the learned distribution of $\beta$.}
\label{fig:beta}
\end{figure*}

In Figure~\ref{fig:beta}, we randomly select 800 sample points and obtained their $\beta$ distributions before training (first row) and after training (second row). The first column represents the distribution of samples with a ground truth index of 1, while the second column represents those with a ground truth index of 3. The red dashed line indicates the ground truth, and the blue dashed line represents the mean of $\beta$ of the samples. Our objective is to bring the blue dashed line closer to the red dashed line.

It can be observed that, compared to the row above, the $\beta$ distributions obtained after training exhibit closer proximity to the ground truth, and they also demonstrate better fitting to Gaussian distributions. This suggests that our model has learned temporal semantics to a certain extent, as the features of samples closer in time are more similar.

\section{More Ablation}
\subsection{Ablation study on the proportion of lateral views}

\begin{table*}[t]
\caption{Ablation study on the proportion of lateral views.}
\label{tab:lateralprop}
\vskip -0.3in
\begin{center}
\begin{tabular}{l|ccccc|c}
\toprule

Lateral proportion (\%)         & Consolidation & Edema & Pl.effusion & Pneumonia & Pneumothorax & \textbf{Avg. Acc.}    \\ \midrule

0       & 55.09\tiny{$\pm$ 1.34}         & 63.32\tiny{$\pm$ 0.78} & 58.06\tiny{$\pm$ 1.40}             &53.58\tiny{$\pm$ 0.40}     & 50.25\tiny{$\pm$ 0.99}        & 58.06\tiny{$\pm$ 0.29}  \\
60       & 55.39\tiny{$\pm$ 1.67}         & 67.32\tiny{$\pm$ 0.88} & 58.63\tiny{$\pm$ 1.00}             &65.98\tiny{$\pm$ 0.91}     & 51.04\tiny{$\pm$ 0.87}        & 59.67\tiny{$\pm$ 0.41}  \\
\rowcolor{gray!20}100           & {60.57}\tiny{$\pm$ 1.18}       &{67.35}\tiny{$\pm$ 0.32} & {58.47}\tiny{$\pm$ 1.50}         & {65.00\tiny{$\pm$ 0.34}}     & {54.18}\tiny{$\pm$ 0.81}     & {61.12}\tiny{$\pm$ 0.34} \\

\bottomrule
\end{tabular}
\end{center}
\end{table*}

Table~\ref{tab:lateralprop} illustrates the results of ablation experiments using different proportions of lateral views. It can be observed that even without the use of lateral views, our mean performance is 58.06, surpassing nearly all baselines. This underscores the effectiveness of local alignment objectives and temporal modeling. As the proportion of lateral views increases, performance also improves, indicating that lateral views provide an additional boost to performance.

\subsection{Ablation study on FMC and RMR}

\begin{table*}[]
\caption{Ablation study on FMC and RMR.}
\vspace{-8pt}
\label{tab:albtemporal}
\vskip -10pt
\begin{center}
\scalebox{0.93}{
\begin{tabular}{lc|cccccc}
\toprule
$\mathcal{L}_{\text{FMC}}$ & $\mathcal{L}_{\text{MRM}}$   & Consolidation & Edema & Pl.effusion & Pneumonia & Pneumothorax & Avg. Acc. \\ \midrule
 \checkmark &  &  53.72 &66.46 &56.93 &63.27 &51.66 &58.33 \\

 & \checkmark  & 53.26 &64.30 &55.02 &65.02 &55.04 &58.53\\
\rowcolor{gray!15}\checkmark & \checkmark  &
60.71 & 67.31 & 58.89 & 65.02 & 55.48 & 61.48 \\
\bottomrule
\end{tabular}
}
\end{center}
\vskip -0.1in
\end{table*}
Table~\ref{tab:albtemporal} presents the ablation experiments conducted on FMC and MRM. The experimental results indicate that MRM contributes to greater benefits, suggesting that more complex regression objectives enable the model to better capture contextual information within sequences. When both are applied, the performance on sequential tasks is optimal, indicating that combining these two objectives, from simple to challenging, allows the model to maximize its understanding of sequence semantics.

\end{document}